\documentclass{article}

\PassOptionsToPackage{square,sort,comma,numbers}{natbib}
\usepackage[final]{neurips_2022}

\usepackage[utf8]{inputenc} % allow utf-8 input
\usepackage[T1]{fontenc}    % use 8-bit T1 fonts
\usepackage{hyperref}       % hyperlinks
\usepackage{url}            % simple URL typesetting
\usepackage{booktabs}       % professional-quality tables
\usepackage{amsfonts}       % blackboard math symbols
\usepackage{nicefrac}       % compact symbols for 1/2, etc.
\usepackage{microtype}      % microtypography
\usepackage{xcolor}         % colors
\usepackage[pdftex]{graphicx}
\usepackage{wrapfig}

\newcommand{\x}{$\times$}
\title{Where Should I Spend My FLOPS? \\ Efficiency Evaluations of Visual Pre-training Methods}

\author{%
Skanda Koppula$^{1}$ \quad Yazhe Li$^{1}$ \quad Evan Shelhamer$^{1}$ \quad Andrew Jaegle$^{1}$ \quad Nikhil Parthasarathy$^{1,2}$ \\
\textbf{Relja Arandjelovic}$^{1}$ \quad \textbf{João Carreira}$^{1}$ \quad \textbf{Olivier Hénaff}$^{1}$ \\
$^{1}$DeepMind \quad $^{2}$NYU Center for Neural Systems\\
\texttt{\{skandak,yazhe,shelhamer,drewjaegle,nikparth,relja,joaoluis,henaff\}@google.com}
}

\begin{document}

\maketitle

\begin{abstract}
Self-supervised methods have achieved remarkable success in transfer learning, often achieving the same or better accuracy than supervised pre-training. Most prior work has done so by increasing pre-training computation by adding complex data augmentation, multiple views, or lengthy training schedules. In this work, we investigate a related, but orthogonal question: given a \textit{fixed} FLOP budget, what are the best datasets, models, and (self-)supervised training methods for obtaining high accuracy on representative visual tasks? Given the availability of large datasets, this setting is often more relevant for both academic and industry labs alike. We examine five large-scale datasets (JFT-300M, ALIGN, ImageNet-1K, ImageNet-21K, and COCO) and six pre-training methods (CLIP, DINO, SimCLR, BYOL, Masked Autoencoding, and supervised). In a like-for-like fashion, we characterize their FLOP and CO$_{2}$ footprints, relative to their accuracy when transferred to a canonical image segmentation task. Our analysis reveals strong disparities in the computational efficiency of pre-training methods and their dependence on dataset quality. In particular, our results call into question the commonly-held assumption that self-supervised methods inherently scale to large, uncurated data. We therefore advocate for (1) paying closer attention to dataset curation and (2) reporting of accuracies in context of the total computational cost.
\end{abstract}

% The past few years have witnessed an explosion of strong self-supervised methods that have achieved remarkable success reaching parity with supervised pre-training.

%Much of prior work has added large numbers of contrastive views, trained for lengthy schedules, and scaled to larger models in order to drive up absolute accuracy on downstream tasks.

% Our results call into question the commonly held hypothesis of the inherent scalability of current self-supervised methods on large-scale uncurated data.

\section{Introduction}

% probably should cite more
Self-supervised learning (SSL) has emerged as the dominant pre-training approach to attain state-of-art performance on complex visual tasks.
The complexity and diversity of SSL methods have exploded in recent years, gradually improving to supersede both supervised pre-training and from-scratch training.
Today's visual transfer learning practitioners have a menagerie of pre-training methods, models, and datasets from which to choose.
Contrastive methods (e.g.\ CLIP \cite{clip}, SimCLR \cite{simclr}, and ReLIC \cite{relic, relicv1}), self-distillation methods (e.g.\ BYOL \cite{byol}, data2vec \cite{data2vec}, DINO \cite{dino}, and ODIN \cite{odin}), and masked autoencoding (MAE) \cite{mae, fastmae} are three of the best-known flavours of modern SSL for visual pre-training.
Despite its strong accuracy on downstream tasks, modern SSL is computationally expensive.
Most methods train for a constant factor more (sometimes an order of magnitude more) \cite{simclr, byol} than their supervised counterparts, necessitate multiple forward passes per gradient step \cite{relic, dino, simclr, byol, odin}, or use expensive Transformer-based decoders that scale quadratically with image resolution \cite{mae}.

In this work, we take a measured look at the computational efficiency of various visual pre-training methods across multiple datasets and model sizes.
The increasing monetary and carbon cost of training large deep networks forces us to ask the question: where should I spend my FLOPs for greatest effect?
Our paper takes the first steps for determining the strongest methods, datasets, and models for pre-training on a fixed budget, as those in academia and industry are both ultimately limited by total computation.
% For example, on the same hardware, pre-training with MAE requires 3\x longer training time (40\% more kg CO$_{2}$) to reach the same downstream image segmentation accuracy as compared to supervised-based pre-training approaches, when using the same uncurated pre-training dataset (JFT-300M) and backbone encoder.
The computational expense of the \mbox{(pre-)training} stage is often ignored, but its performance has material impact: (1) faster pre-training reduces iteration time for research, (2) less computationally-expensive methods are accessible to more research groups, (3) with larger data and models becoming more accessible, compute becomes the bottleneck, and training efficiency gains therefore translate into greater performance gains \cite{chinchilla} (4) these improvements compound as many modalities are combined during pre-training \cite{flamingo, wang2022image}, and (5) efficiency is simply better for the environment. Lastly, accounting for computation is more empirically rigorous: more computation often yields better accuracy (more steps, views, etc.), and just as scaling laws for models show improvement with more parameters \cite{chinchilla}, reporting computation-normalized performance enables more like-for-like comparisons.

% Paragraph on prior work
% pretty common to see model size vs.\ performance
% BiT paper has some nice transfer: many large ResNet models didn't 
% ViTs supervised also scale nicely --> comment on?
Only a few prior works have compared pre-training datasets across SSL methods \cite{bit, vit, i21kmasses, xie2022data}. 
Notably, SEER \cite{seer} presents strong results pre-training on the large-scale uncurated IG dataset, but do not compare pre-training datasets while controlling for others factors.
% across methods but not datasets, with its focus on on the large, uncurated IG dataset.
Additionally, while model scaling curves are popular in the SSL literature \cite{byol, relic, seer}, backbone model size is not equivalent to computational efficiency, as other aspects like the number of steps can dominate overall FLOP cost. 
We have yet to find prior work comparing (or reporting) effectiveness as a function of FLOPs.
We address this gap by profiling four popular self-supervised methods (BYOL \cite{byol}, SimCLR \cite{simclr}, DINO \cite{dino}, and MAE \cite{mae}) and two supervised methods (CLIP \cite{clip} and standard classification). We compute their per gradient step FLOP cost (post-compilation, running on the same accelerator hardware), and use this to cross-compare the three axes of pre-training method, model size (ViT-B vs.\ ViT-L \cite{vit}), and dataset choice (COCO, ImageNet 1K, ImageNet-21K, ALIGN, and JFT-300M).

% perhaps define pareto optimal
We find that the computational efficiency of SSL methods differs by up to an order magnitude, and that FLOP-efficient methods can train significantly larger backbones than FLOP-inefficient ones, while being cheaper to run. % FLOP-expensive methods can exceed the cost of training efficient methods with . 
The best efficiency/accuracy trade-off (that is, the pareto-optimal curve) of CLIP dominates that of classification supervised methods on large-scale datasets, nearly matched or exceeded by MAE, depending on the dataset and downstream task. Below all these methods are SimCLR, BYOL, and DINO, in that order.
Dataset quality and curation level also matter significantly: when controlling for FLOPs, most methods perform significantly worse on uncurated data. 
While SSL offers the promise of learning effectively on in-the-wild data, most methods are tuned explicitly for, or simply perform better on, the highly-curated ImageNet-1K and 21K datasets \cite{imagenet1k}.

We hope this work sparks future research into visual SSL methods that learn more effectively and scalably on uncurated data, following the initial promise of \cite{seer}.
As we progress towards richer media and larger datasets, the need for efficient SSL algorithms will grow, and we hope to encourage methods that achieve the promise of label-efficient and computationally scalable learning.

\vspace{-3mm}
\section{Methods}
\vspace{-3mm}
\subsection{Pre-training Methods and Models}
\label{sec:pre-training_methods}
\vspace{-2mm}
We run four common self-supervised methods (BYOL \cite{byol}, SimCLR \cite{simclr}, DINO \cite{dino}, and MAE \cite{mae}) and two supervised methods (CLIP \cite{clip} and standard softmax classification). These methods span the contrastive, self-distillation, and reconstruction-based method families popular in modern SSL. These methods are pre-trained with four computational budgets (from 62,500 to 750,000 steps), with a batch size of 4096 (except for DINO, which is run with a batch size of 1024 and for 4\x \ longer). For context, 250,000 steps at batch size 4096 is equivalent to seeing the same number of training images as in the commonly used 800-epochs of ImageNet-1K training in SSL. %The methods are implemented in JAX~\cite{jax}, with original implementations or reproductions that match the original ImageNet-1K pre-training and finetuning classification accuracies. 
We use two sizes of vision transformer as the backbone vision encoder: ViT-B/16 (87M parameters) and ViT-L/16 (305M parameters). More detail on pre-training and optimization details is in Section~\ref{sec:pre-training_methods_appendix}.

Because different masking ratios in MAE affect both accuracy and FLOP cost, we sweep across four different masking ratios (55\%, 65\%, 75\%, and 85\%) in the neighborhood of the default 75\%. Because the decoder in MAE can occupy up to 85\% of the total per-step time and FLOP cost (Section~\ref{sec:mae_profile_appendix}), we experiment with both the default 8-layer MAE decoder, and a `tiny' one-layer decoder that enables 72\% faster MAE training. Our CLIP model includes two sizes of a BERT-style language model: either a `mini` 4-layer, 256D-hidden encoder (12M parameters), or a `medium` 8-layer, 512D-hidden encoder (44M parameters). This setup mimics the architectures used in \cite{flamingo}. To train on the hierachical classification-based JFT, we use the softmax-based averaging method described in Section~\ref{sec:pre-training_methods_appendix}. For CLIP, we use a pseudo-caption passed to the language encoder of the form `\texttt{This is a photo of <label> and <label2> ... <label\_n>.}'.
% \vspace{-3mm}
\subsection{Datasets}
\vspace{-3mm}
We use five pre-training datasets: COCO (118K images), ImageNet-1K (1.2M images, `I1K'), ImageNet-21K (15M images, `I21K', a superset of ImageNet-1K), JFT-300M (302M images), and ALIGN (1.6B images). These datasets span both size and amounts of curation. I1K possesses human-guided views typically framed around a centered object, with 1000 hand-picked classes. ALIGN and JFT are two mildly-filtered web-scale datasets.
% \vspace{-3mm}
\subsection{Measurement and Evaluation}
\vspace{-2mm}
We evaluate downstream performance by finetuning the pre-trained encoders for semantic segmentation on the popular dataset, ADE20K \cite{ade20k}. In our experiments, we find this task to be reasonably representative of other complex visual tasks, such as object detection (\ref{sec:object_detection}), but also demand more spatial understanding from representations as compared to more simple objectives such as ImageNet-1K classification. We use the same finetuning segmentation head as in \cite{byol, detcon, odin, moco}, which takes as input the reshaped last layer tokens from the ViT encoder. We finetune each model for a fixed 80 epochs similar to a standard ADE20K setup \cite{mmseg_ade20k_80epochschedule}. More details can be found in Section~\ref{sec:finetuning_appendix}.

We use a profiling tool for JAX to trace and count the floating point operations of a single pre-training step, for  each pre-training method and model size. FLOP measurements occur after model code is lowered and compiled to a specific accelerator setup (8 trays of TPUs \cite{tpu}). This provides us with an accurate FLOP estimate that correlates well with seconds per training step (Section~\ref{sec:train_length_estimation_appendix}). CO$_2$ emissions are estimated by translating the total FLOP count to hypothetical training time on a 32 TPUv3 group. From this, power consumption and carbon emissions are estimated based on the accelerator power draw and datacenter location and efficiency using an online tool \cite{co2_impact}.
% \vspace{-4mm}
\section{Results}
\vspace{-3mm}
\subsection{Compute Efficiency of Supervised and Self-supervised Learning}
\vspace{-5mm}
\begin{figure}[h]
  \centering
  \hbox{
  \hspace{-12mm}
  \includegraphics[width=1.2\linewidth]{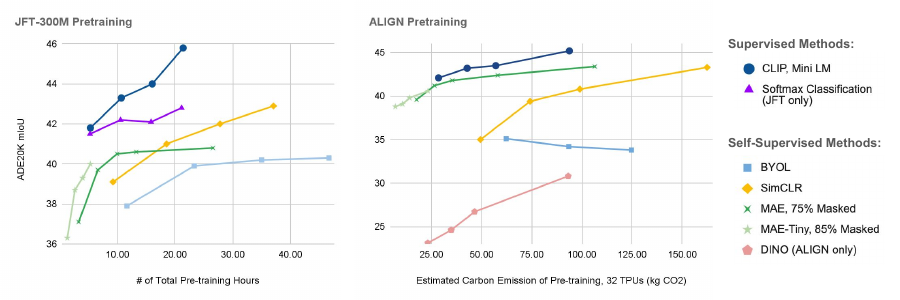}
  }
  \vspace{-4mm}
  \caption{Segmentation accuracy after pre-training ViT-B models on ALIGN or JFT. Supervision consistently outperforms self-supervision in both accuracy (up) and efficiency (left) for this image segmentation evaluation, with MAE closely following supervised pre-training. Best viewed in color.}
  \label{fig:jft-align-efficiency}
  % \vspace{-2mm}
\end{figure}

% We pre-train ViT-B/16 models on JFT-300M using our seven supervised and self-supervised methods with four fixed cosine LR schedules: a pre-training schedule for 62500, 125000, 187500, and 250000 steps. On the larger ALIGN, we train each model for either 125000, 187500, 250000, and 410000 steps. The downstream results of each of these models are shown as points in Figure~\ref{fig:jft-align-efficiency}. In order to maximize regions of iso-FLOP overlap, since MAE is particularly FLOP-efficient, we pre-train two additional MAE models, one for 500000 steps on JFT and the other for 750000 steps on ALIGN. These are as the rightmost points on the green curves in Figure~\ref{fig:jft-align-efficiency}.
\vspace{-2mm}
Our first set of accuracy-efficiency curves are shown in Figure~\ref{fig:jft-align-efficiency}. Self-supervised methods are generally less FLOP efficient and supervised representations dominate the efficiency pareto-front. In particular, CLIP pre-training demonstrates strong compute scaling and outperforms the standard classification objective. BYOL performs poorly on these benchmarks and we hypothesize that this is due to the `strong' normalization required by BYOL \cite{byolbatchstats} (which is not provided in a standard ViT backbone). While SimCLR, BYOL, and DINO are substantially less efficient than supervised learning, MAE closely follows the pareto-efficiency front of the supervised methods, and in our object detection results (\ref{sec:object_detection}), MAE even slightly exceeds to supervised curves.
\vspace{-1mm}
\subsection{Scaling Model Size or Training Length?}
\vspace{-2mm}
\begin{figure}[h]
  \centering
  \hbox{
  \hspace{-15mm}
  \includegraphics[width=1.2\linewidth]{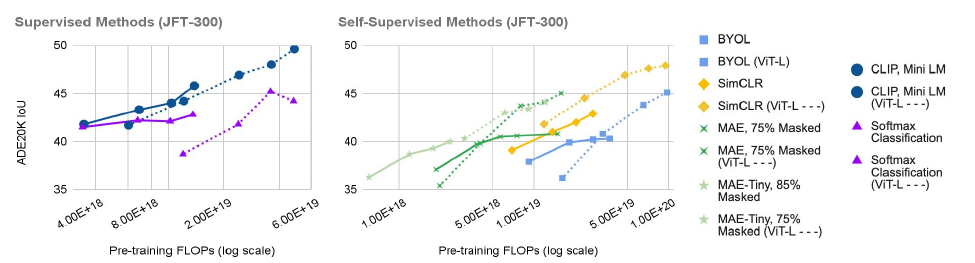}
  }
  \vspace{-2mm}
  \caption{Segmentation accuracy-efficiency curves for JFT pre-training with ViT-B (solid) and ViT-L (dashed). Up is more accurate and left is more efficient. Best viewed in color.}
  \label{fig:jft-largemodel-efficiency}
  \vspace{-3mm}
\end{figure}
In our second set of accuracy-efficiency curves, Figure~\ref{fig:jft-largemodel-efficiency}, we show the compute scaling of ViT-B and ViT-L which contains roughly 3\x more parameters. For a given pre-training  budget, it is not always beneficial to scale the model size to obtain better accuracy: for example, with CLIP, a ViT-B trained with $3 \times 10^{18}$ FLOPS outperforms a ViT-L trained with $6 \times 10^{18}$ (a 3\x gain in efficiency). More generally, ViT-B's FLOP-efficiency exceeds that of ViT-L across the computational regime we tested.

% and a $8\times 10^{18}$ FLOP budget (i.e.\ approximately 15 hours on 32 TPUv3s), it is better to use a ViT-B trained for <100K steps, than to jump to a ViT-L with any training length.

The story is slightly different for self-supervised methods: larger models tend to have an on-par or better compute/accuracy curve for the FLOP regimes tested. This is similar to what is observed in language model scaling \cite{kaplan2020scaling, chinchilla}. For most methods, the small and large model curves intersect, indicating the point at which it is better to switch to larger model sizes for a given FLOP budget.
\vspace{-1mm}
\subsection{Dataset Choice and Curation}
\vspace{-2mm}
We also evaluate our SSL methods across multiple pre-training datasets, as described in Sections \ref{sec:pre-training_methods} and \ref{sec:pre-training_methods_appendix}. Downstream results for our ViT-B/16 models are shown in Table~\ref{tab:datasets}. Only BYOL (pre-trained on ImageNet-21K)  
outperforms ImageNet-1K pre-training. We suspect the power of I1K pre-training stems from a combination of (1) I1K's more even class labels and label curation, (2) implicit human supervision (centering of the photos, framing the object) \cite{dnc}, and (3) tuning methods around the I1K pre-training benchmark (e.g.\ in choosing augmentations) \cite{randaugment}.

Only SimCLR demonstrates consistent improvement as dataset size increases (apart from I1K). For other methods, such as BYOL and DINO, the best results are not with the largest pre-training dataset, ALIGN. Furthermore, ALIGN (which has mild pre-processing, a caption frequency-based filtering step) yields consistently better results than less curated (albeit smaller) JFT. We had difficulty stabilizing long-schedule training for DINO on JFT, longer than the listed 500K steps @ 1K batch size. Our efforts tuning gradient clipping, decay, and learning rate were unsuccessful.

\begin{table}[b]
  \vspace{-3mm}
  \centering
  \begin{tabular}{llllll}
    \toprule
    Method      & \multicolumn{5}{c}{Pre-training Dataset} \\
    \midrule
    \      & ImageNet-1K     & COCO & ImageNet-21K+1K & JFT & ALIGN\\
    \midrule
    BYOL   & 41.7 & {37.1} & \textbf{42.4} & {40.3} & {35.1} \\
    SimCLR & \textbf{44.0} & {38.9} & {42.9} & {42.9} & {43.3} \\
    MAE    & \textbf{42.3} & {41.0} & {40.5} & {40.6} & {42.3} \\
    DINO   & \textbf{36.7} & {31.4} & {34.2} & 29.2* & {32.2} \\
    \bottomrule
  \end{tabular}
  \vspace{2mm}
  \caption{ADE20K finetuning results (mIoU) after pre-training on various datasets with four SSL methods (fixed 250K steps). Bold indicates the best pre-training dataset for that method. DINO training with long schedules on JFT (marked with an asterisk) exhibits significant training instability.}
  \label{tab:datasets}
\end{table}

% \begin{figure}
%   \centering
%   \fbox{\rule[-.5cm]{0cm}{4cm} \rule[-.5cm]{4cm}{0cm}}
%   \caption{Sample figure caption.}
% \end{figure}

% \subsection{Tables}

% \begin{table}
%   \caption{Sample table title}
%   \label{sample-table}
%   \centering
%   \begin{tabular}{lll}
%     \toprule
%     \multicolumn{2}{c}{Part}                   \\
%     \cmidrule(r){1-2}
%     Name     & Description     & Size ($\mu$m) \\
%     \midrule
%     Dendrite & Input terminal  & $\sim$100     \\
%     Axon     & Output terminal & $\sim$10      \\
%     Soma     & Cell body       & up to $10^6$  \\
%     \bottomrule
%   \end{tabular}
% \end{table}
% \vspace{-3mm}
\section{Conclusion}
\vspace{-3mm}
We take a first step towards more rigorously characterizing the computational efficiency of modern supervised and self-supervised pre-training methods along the axes of pre-training methods, dataset, and model size. We advocate that for sake of research accessibility, environmental impact, and robust comparisons, more attention be paid to the computational efficiency of visual pre-training methods. Future directions include more fine-grained characterization of our efficiency curves into power scaling laws, design of new SSL objectives and curation methods that overcome dataset sensitivity, and methods that push the learning efficiency of SSL to be competitive with supervised pre-training.

% Most of the margin problems come from figures positioned by hand using
% \verb+\special+ or other commands. We suggest using the command
% \verb+\includegraphics+ from the \verb+graphicx+ package. Always specify the
% figure width as a multiple of the line width as in the example below:
% \begin{verbatim}
%   \usepackage[pdftex]{graphicx} ...
%   \includegraphics[width=0.8\linewidth]{myfile.pdf}
% \end{verbatim}
% See Section 4.4 in the graphics bundle documentation
% (\url{http://mirrors.ctan.org/macros/latex/required/graphics/grfguide.pdf})

% A number of width problems arise when \LaTeX{} cannot properly hyphenate a
% line. Please give LaTeX hyphenation hints using the \verb+\-+ command when
% necessary.

\newpage
\newpage

%%%%%%%%%%%%%%%%%%%%%%%%%%%%%%%%%%%%%%%%%%%%%%%%%%%%%%%%%%%%%%%%%%%%%%%%%%%%%%%
%%%%%%%%%%%%%%%%%%%%%%%%%%%%%%%%%%%%%%%%%%%%%%%%%%%%%%%%%%%%%%%%%%%%%%%%%%%%%%%

\bibliographystyle{plain} % We choose the "plain" reference style
\bibliography{refs} % Entries are in the refs.bib file

%%%%%%%%%%%%%%%%%%%%%%%%%%%%%%%%%%%%%%%%%%%%%%%%%%%%%%%%%%%%%%%%%%%%%%%%%%%%%%%
%%%%%%%%%%%%%%%%%%%%%%%%%%%%%%%%%%%%%%%%%%%%%%%%%%%%%%%%%%%%%%%%%%%%%%%%%%%%%%%

% \newpage

\appendix

\section{Appendix}

\subsection{CO$_2$ estimation}
\label{sec:co2_estimation_appendix}

To estimate CO$_2$ emissions of pre-training, we first must estimate total training time on some fixed hardware (32 TPUv3s, in our case). Unfortunately, total training time is not something that typically can be measured in an apples-to-apples fashion, because each of our pre-training jobs runs on different hardware (various flavours and counts of TPUs and GPUs), based on availability. So, we follow a three step process to obtain more accurate estimates on standardized hardware:
\begin{enumerate}
    \item We measure the total per step FLOP cost of each pre-training job, through a hardware-targeted JAX profiler.
    \item We convert this FLOP cost to the steps per second on 32 TPUv3s. This is done by collecting steps/second measurements from short-schedule pre-training jobs across all pre-training methods. From this, we're able to model the power-law correlation between per step FLOP cost with steps per second on TPUv3 hardware ($R^2=0.97$), and translate more easily between FLOP cost and estimated steps per second. This also has the added benefit of smoothing the small performance measurement variability between runs.
    \item With the steps per second of a training job, and the training length (in steps), we calculate the total estimated training on the target hardware topology.
\end{enumerate}
Given the total training time, we use the online calculator from \cite{co2_impact}, to compute an estimated equivalent power and carbon cost for a single TPUv3 chip in the \texttt{europe-west1} region. We scale this by 32 (using 32 TPUs in our estimate) to obtain the power and carbon cost numbers we report in the paper.

We note, however, that the datacenters in which our experiments are run are completely carbon offset.

\subsection{FLOP estimation}
\label{sec:train_length_estimation_appendix}

Post-compiled FLOPs can slightly differ from estimated FLOPs, as the compiler can excise or expand parts of the computation graphs, and/or leverage hardware-specific specific SIMD compute instruction ops, based on characteristics of the target hardware. Hence, rather than numerically adding up the theoretical FLOP count based on the architecture implementation, we use AX's ability to trace through the lowered, post-compilation code to measure the FLOP count.

\begin{center}
\begin{figure}[h]
\centering
    \vspace{-6mm}
    \includegraphics[width=0.8\linewidth]{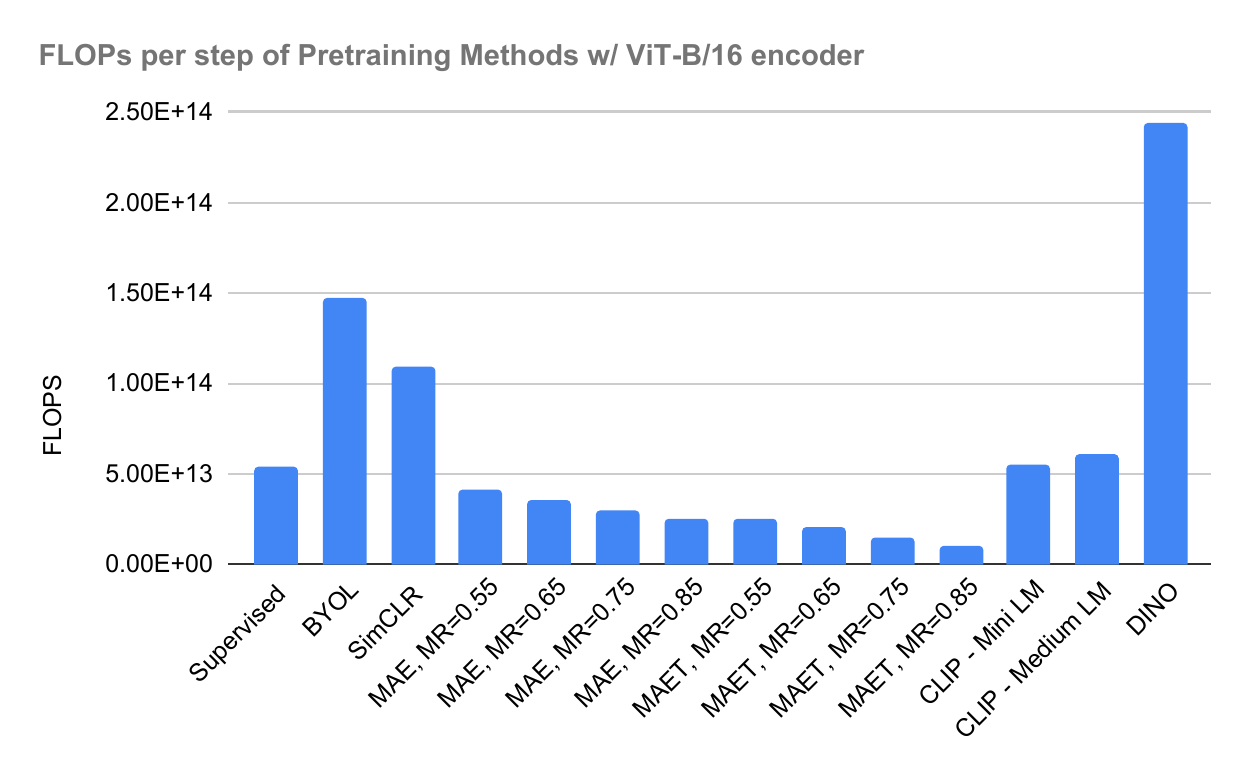}
\label{fig:flop-cost-methods}
\caption{Per step FLOP count of various supervised and self-supervised pre-training methods.}
\vspace{-5mm}
\end{figure}
\end{center}

Figure~\ref{fig:flop-cost-methods} reports our measured FLOP cost per training step for the evaluated pre-training methods.

\vspace{-10mm}
\subsection{Pre-training configuration}
\label{sec:pre-training_methods_appendix}

To gather a wide sample of points along FLOP/accuracy curve, we train all pre-training methods for a varying number of steps. Each pre-training length needs to be trained separately to fit with the optimal cosine learning rate schedule that warms up and decays back to zero at the end of the pre-training length. On JFT, we pre-train the all methods except DINO for 62500, 125000, 187500, and 250000 steps at batch size 4096. To match the official implementation, we train DINO with a 1024 batch size, for 125000, 250000, and 500000 steps. To obtain an additional data point further along the curve, in a higher FLOP regime, we train an MAE for 500K steps as well. On ALIGN, we pre-train all methods except DINO for 125000, 187500, 250000, and 410000 steps at batch size 4096. To match the official implementation, we train DINO with a 1024 batch size, for 125000, 250000, 500000, and 750000 steps. To obtain an additional data point along the curve, we train an MAE for 750K steps as well. On I21K, I1K, and COCO, we train all methods for 250K at batch size 4096.

For all methods, we apply standard ImageNet image normalization for both pre-training and finetuning \cite{i1k_pixel_norm}, with image size of (224, 224). We use the MAE ViT \cite{mae} as the backbone in SimCLR, BYOL, and CLIP (using e.g., the cosine position encodings). Supervised classification pre-training uses the backbone from \cite{vit}, and DINO uses the standard \cite{dino} ViT implementation.

For training softmax classification on the hierarchical JFT, we use the multi-label one-hot averaging scheme from \cite{nfnet}. This scheme performs similarly to a sigmoid-based classification. In pseudo-code:

\begin{verbatim}
    y = one_hot(jft_labels, num_jft_classes)
    y = np.sum(y, -2)
    y = y / np.sum(y, -1, keepdims=True)
    y = smooth_labels(y, 0.1)
    classification_loss = softmax_cross_entropy(logits, y, reduction=`mean')
\end{verbatim}

For CLIP and softmax classification methods, we set the maximum number of allowable labels in JFT to 8. For images with more labels than this, the label list will be truncated; for images with fewer labels, it will be padded.
 
For CLIP, we re-use the same training crop and rescaling augmentations as \cite{clip} and the contrastive pre-training for \cite{flamingo}. For all datasets, we use AdamW with weight decay of 2e-2, b1=0.9, b2=0.95, and a gradient clipping thresholding the global norm to 1.0. We use warm-up cosine decay learning rate schedule, with base learning rate of 1e-3, starting learning rate of 1e-7, and 10K warmup steps. The joint embedding space has dimension 768 and the language encoder is fed a maximum caption length of 55 characters.

 For BYOL, we re-use the same training crop and rescaling augmentations as \cite{byol} and the official I1K implementation \cite{byol_implementation}. For JFT and ALIGN, we use a target EMA of 0.99, base learning rate of 3e-4, and weight decay of 1e-1. For I1K and I21K, we use a target EMA of .996 and weight decay of 1e-1. For COCO, we use a target EMA of .97, weight decay of 1e-6, and learning rate of 3e-4. For all datasets, we use the Adam optimizer with {b1=0.9, b2=0.999}. We use a warmup cosine schedule, with warmup steps equal to 5\% of the total steps in the entire schedule. Our SimCLR implementation uses these same optimization settings (but no target EMA).
 
 For DINO, we use the same augmentations as in the official implementation \cite{dino_implementation}. We use the Adam optimizer with b1=0.9, b2=0.999. For most datasets, we use gradient clipping of 1.0 (although this is lowered for larger more unstable datasets like JFT to down to 0.1), 10 local views, and a cosine LR schedule. Our weight decay starts at 0.04 and ends at 0.4. We use an LR of 0.00075 for most datasets, except for JFT long schedules where most of our best results and most stable training is with an LR of 0.0001. In the LR schedule, we use a warmup of 12K steps and minimum LR of 2e-6.
 
 For classification-based supervised ViT training, we use the standard crop-resize augmentation as is standard in I1K training and drop path rate of 0.1. We use a cosine schedule for the learning rate, with a base value LR of 1e-4, and 6K warmup steps. We use a weight decay of 0.3. We use the Adam optimizer with b1=0.9, b2=0.95. These settings are the same across datasets. We use gradient clipping thresholding the global norm at 1.0. Importantly, we use a common initialization trick of initializing values to keep the initial range within 0.02, and initializing the head bias to -10 \cite{vit, deit}.
 
 For all datasets, we use AdamW with weight decay of 2e-2, b1=0.9, b2=0.95, and a gradient clipping thresholding the global norm to 1.0. We use warm-up cosine decay learning rate schedule, with base learning rate of 1e-3, starting learning rate of 1e-7, and 10K warmup steps. The joint embedding space has dimension 768 and the language encoder is fed a maximum caption length of 55 characters.
 \vspace{-6mm}
\subsection{Fine-tuning configuration}
\label{sec:finetuning_appendix}

We finetune all models for a fixed 80 epochs on ADE20K, using 16 TPUv1s and a per device batch size of 2. For both training and evaluation, we use an image size of [512, 512]. We add photometric augmentations and randomly rescaled image cropping (minimum scale factor of 0.5, maximum of 2.0, sampling the scaling factor in steps of 0.25) during training. For photometric augmentations, we use the color augmentation function in \cite{byol_color_aug}, with settings \texttt{brightness=0.1, contrast=0.1, saturation=0.1, hue=0.1, apply\_prob=0.5}. For evaluation, we take a center crop of the desired size. We use the \texttt{train} split for finetuning and report numbers on the \texttt{val} split.

We use nearly the same optimization settings as \cite{mmseg_ade20k_80epochschedule}. We use a warmup+polynomial LR schedule, with 1500 warmup steps, and the same configuration (\texttt{lr\_config}) as \cite{mmseg_ade20k_80epochschedule}. We use ADAM optimizer with \texttt{b1=0.9, b2=0.999, eps=1e-8} and constant weight decay. The segmentation head on top of the ViT encoder takes as input the last layer ViT tokens, reshaped into a square grid, and has the same architecture as implemented in \cite{mmseg_fcn_head} (two 3\x3 convolutional layers, and channel dimension of 256). We apply gradient clipping with global norm cutoff of 1.

Similar to \cite{mae}, we ran an exploratory learning rate (LR) search across \{1e-5, 5e-5, 1e-4, 5e-4\} and weight decay (WD) values \{0.1, 0.05, 0.01\} on the 250K-step pre-trained models. We used the best performing values as the fine-tuning settings for that method and dataset. Generally, we found \texttt{LR=1e-4, WD=0.05} as the best setting for MAE pre-trained networks and \texttt{LR=5e-5, WD=0.1} as optimal for BYOL, CLIP, and SimCLR pre-trained networks. For DINO, the optimal settings were generally \texttt{LR=5e-05, WD=0.05}. To obtain final numbers, we ran five fine-tuning jobs  with different seeds using the optimal LR/WD settings and reported the median.

From our experiments, this same ViT+segmentation head architecture trained from scratch on this benchmark yields 20.8 mIoU, worse than all supervised or self-supervised based initialization.
\vspace{-2mm}
\subsection{Profiling of MAE}
\label{sec:mae_profile_appendix}

The 75\% masking ratio that is commonly used for image tasks is not always optimal across datasets and MAE variants: e.g. on COCO, find that 65\% masking to reach highest absolute downstream accuracy. For all other datasets, we find that 75\% is optimal. On ALIGN, 85\% masking also results in similar performance. With the one-layer decoder, we find that slightly higher masking rates also sometimes have better absolute downstream accuracy and efficiency-normalized accuracy. In Figures~\ref{fig:jft-align-efficiency} and \ref{fig:jft-largemodel-efficiency}, we report masking ratios that result in the highest efficiency/accuracy curve (that envelopes the other ratio's curves, by being further to the left and higher). For Table~\ref{tab:datasets}, we report results for the masking ratios that result in highest absolute downstream accuracy.

By masking a significant portion of input tokens, masked autoencoding has a particularly lightweight encoder. From our profiling, we find that the wall-clock performance improvements a over a baseline ViT encoder is often significantly lower than the masking ratio, because of the computational overhead of the decoder. For example, with an 85\% masking ratio, wall-clock time savings are slightly under 30\%, because computation becomes dominated by the decoder. For example, with an 75\% masking ratio, roughly 65\% of the time to compute a forward and backward pass is occupied by operations for the decoder. With a 55\% masking ratio, this decoder percent falls to roughly 45\%. This motivated our use of the `tiny` MAE variant, with a simple one-layer decoder. For a 75\% masking ratio, the wall-clock speed of tiny-MAE is 95\% faster.

\subsection{Transfer to Object Detection}
\label{sec:object_detection}

To verify the robustness of our overall conclusions, in addition to semantic segmentation, we also test our pre-trained representations on the canonical COCO-2017 \cite{coco} object detection task. For finetuning on COCO, we use a Mask-RCNN architecture with a modified ViT-B/16 backbone (instead of a ResNet-50 backbone), similar to \cite{vitdet}. The Mask-RCNN has the same configuration and optimization settings as the ResNet-50 based Mask-RCNN in \cite{detcon, odin}, finetuning for 22K steps. We slightly modify the \cite{vitdet} architecture for our evaluations, and remove their windowed attention and extra convolutional pooling operations inserted into each transformer block (and instead use the ViT backbone unchanged). This reduces the computation efficiency during finetuning, but enables us to use the same exact pre-trained backbone with little modification to the attention mechanism/masking. We report \texttt{Box AP@[IoU=0.50:0.95|area=all|maxDets=100]} in our results. For reference, from scratch training for the same number of steps obtains 22.0 mAP.

Table~\ref{tab:object_detection_datasets} reports COCO object detection results of SSL pre-training on various datasets. Similar to our segmentation results, we find that ImageNet-1K pre-training often outperforms much larger pre-training datasets, with exception of the "full ImageNet" (I21K+I1K), where the performance improvement over I1K alone is small ($<=0.5$ mAP). More broadly, we find in this downstream task as well, that pre-training dataset size does not correlate with downstream performance.
\begin{table}[t]
  \vspace{-3mm}
  \centering
  \begin{tabular}{llllll}
    \toprule
    Method      & \multicolumn{5}{c}{Pre-training Dataset} \\
    \midrule
    \      & ImageNet-1K   & COCO   & ImageNet-21K+1K & JFT    & ALIGN\\
    \midrule
    BYOL   & 40.0          & {39.2} & \textbf{40.3}   & {39.1} & {27.9} \\
    SimCLR & \textbf{42.0} & {41.7} & {41.2}          & {40.3} & {35.6} \\
    MAE    & \textbf{44.8} & {44.6} & {43.1}          & {42.6} & {44.2} \\
    DINO   & 32.8          & {29.8} & \textbf{33.2}   & 30.9*  & {32.3} \\
    \bottomrule
  \end{tabular}
  \vspace{2mm}
  \caption{COCO finetuning results (mAP) after pre-training on various datasets with four SSL methods. Bold indicates the best pre-training dataset for that method. DINO pre-training with long schedules on JFT (marked with an asterisk) exhibited some training instability.}
  \label{tab:object_detection_datasets}
  \vspace{-7mm}
\end{table}

\begin{center}
\begin{figure}[h]
\centering
    \vspace{-4mm}
    \includegraphics[width=0.8\linewidth]{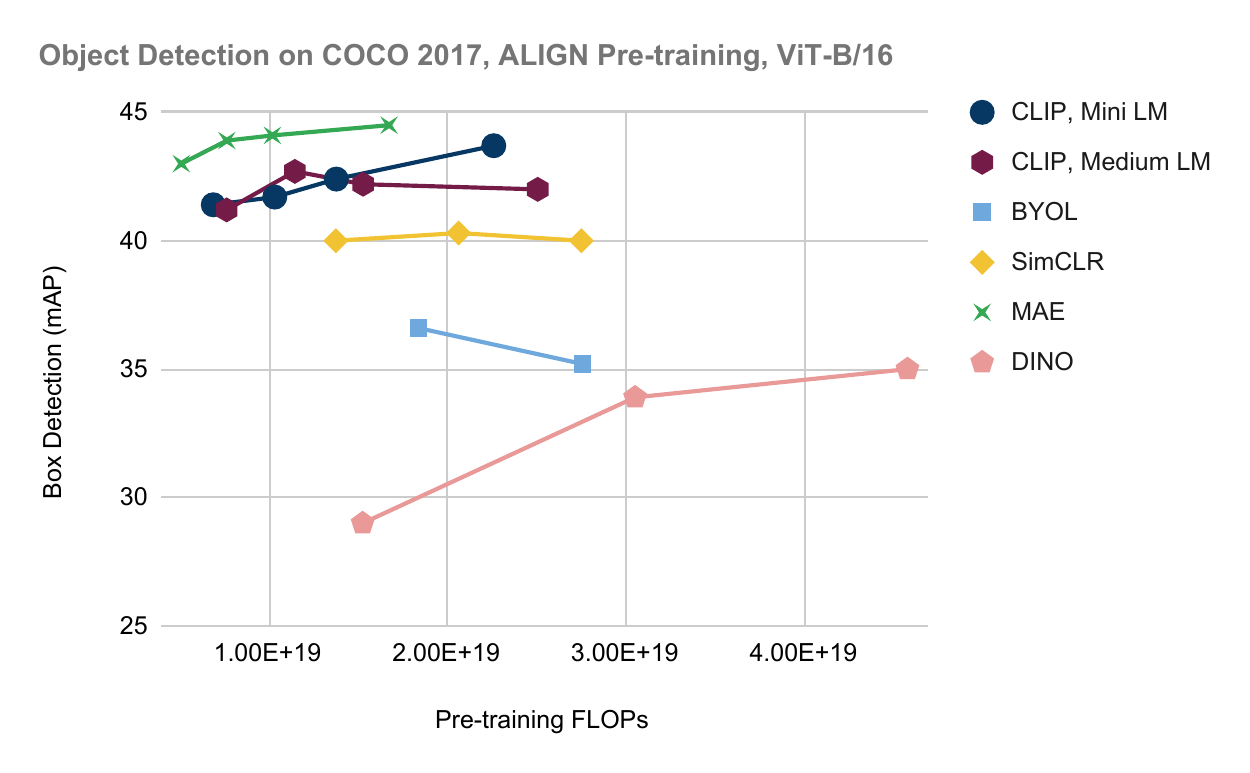}
\label{fig:detection-efficiency-align}
\caption{Object detection performance after pre-training ViT-B models on ALIGN. MAE outperforms CLIP-based supervision which in turn outperforms other self-supervision methods in both accuracy (up) and efficiency (left). Best viewed in color.}
\end{figure}
\end{center}

\begin{center}
\begin{figure}[h]
\centering
    \vspace{-6mm}
    \includegraphics[width=0.8\linewidth]{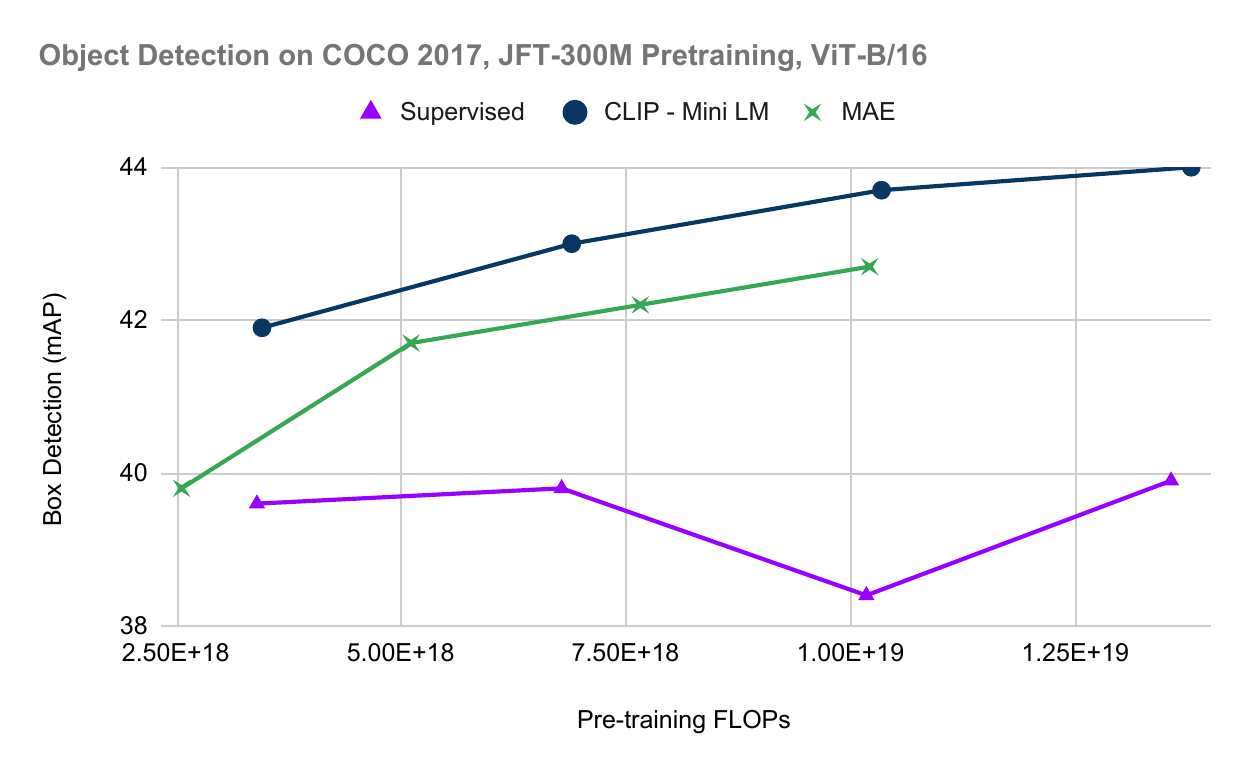}
\label{fig:detection-efficiency-jft}
\caption{Object detection performance after pre-training ViT-B models on JFT-300M. CLIP outperforms MAE which outperforms standard classification supervision. Best viewed in color.}
\end{figure}
\end{center}

Figure~\ref{fig:detection-efficiency-align}.4 and \ref{fig:detection-efficiency-jft}.5 shows the FLOP efficiency of each method, measured on the object detection task. We obtain similar results to segmentation, with the exception that MAE overtakes one or two supervised methods, depending on the pre-training dataset. MAE performs strongly both in absolute terms, and with respect to its compute efficiency.

\end{document}